# WaveletFormerNet: A Transformer-based Wavelet Network for Real-world Non-homogeneous and Dense Fog Removal

Shengli Zhang[a,1], Zhiyong Tao[a,**,2] and Sen Lin[b,3]

[a]*School of Electronic and Information Engineering, Liaoning Technical University, Huludao, 125105, Liaoning, China*
[b]*School of Automation and Electrical Engineering, Shenyang Ligong University, Shenyang, 110159, Liaoning, China*

ABSTRACT

Although deep convolutional neural networks have achieved remarkable success in removing synthetic fog, it is essential to be able to process images taken in complex foggy conditions, such as dense or non-homogeneous fog, in the real world. However, the haze distribution in the real world is complex, and downsampling can lead to color distortion or loss of detail in the output results as the resolution of a feature map or image resolution decreases. Moreover, the over-stacking of convolutional blocks might increase the model complexity. In addition to the challenges of obtaining sufficient training data, overfitting can also arise in deep learning techniques for foggy image processing, which can limit the generalization abilities of the model, posing challenges for its practical applications in real-world scenarios. Considering these issues, this paper proposes a Transformer-based wavelet network (WaveletFormerNet) for real-world foggy image recovery. We embed the discrete wavelet transform into the Vision Transformer by proposing the WaveletFormer and IWaveletFormer blocks, aiming to alleviate texture detail loss and color distortion in the image due to downsampling. We introduce parallel convolution in the Transformer block, which allows for the capture of multi-frequency information in a lightweight mechanism. Such a structure reduces computational expenses and improves the effectiveness of the network. Additionally, we have implemented a feature aggregation module (FAM) to maintain image resolution and enhance the feature extraction capacity of our model, further contributing to its impressive performance in real-world foggy image recovery tasks. Through extensive experiments on real-world fog datasets, we have demonstrated that our WaveletFormerNet achieves superior performance compared to state-of-the-art methods, as shown through quantitative and qualitative evaluations of minor model complexity. Additionally, our satisfactory results on real-world dust removal and application tests showcase the superior generalization ability and improved performance of WaveletFormerNet in computer vision-related applications compared to existing state-of-the-art methods, further confirming our proposed approach's effectiveness and robustness.

## 1. Introduction

Haze is a common atmospheric phenomenon that causes distortion and degradation of images. Image dehazing techniques are significant for many computer vision tasks, such as remote sensing processing [23] and video analysis and recognition [39, 49]. In recent years, image dehazing has been a hot research topic in computer vision and image processing, serving as an essential low-level image recovery task and a preprocessing step for high-level vision tasks.

Many previous dehazing methods [21, 17, 56, 6, 38, 54] have used the classical atmosphere scattering model (ASM) [32, 36, 35] to characterize the degradation process of hazy images by Eq. (1):

$$I(x) = J(x)t(x) + A(1 - t(x)) \tag{1}$$

where $I(x)$ and $J(x)$ are the degraded images and the clear images, respectively. $A$ represents global atmospheric light; $t(x) = e^{-\beta d(x)}$ is the transmission map, where $\beta$ and $d(x)$ represent atmospheric scattering parameters and scene depth, respectively. The main idea of early prior-based dehazing methods is to estimate the medium transmission map $t(x)$ and the global atmospheric light $A$ by handcraft; these methods have made significant progress.

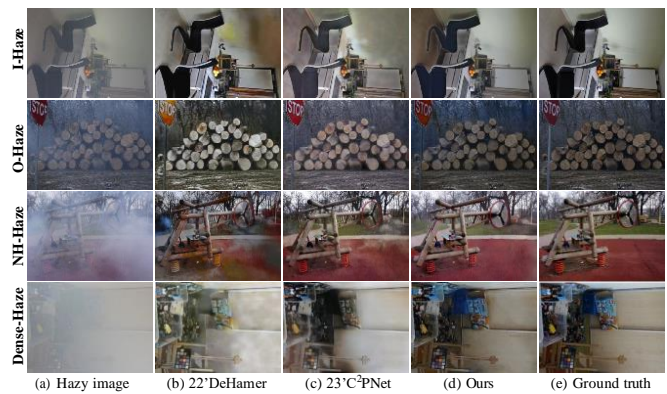

(a) Hazy image  (b) 22'DeHamer  (c) 23'C²PNet  (d) Ours  (e) Ground truth

**Figure 1:** Qualitative dehazing performance comparison among WaveletFormerNet and SOTA methods on four benchmark real-world hazy datasets (I-Haze, O-Haze, NH-Haze, and Dense-Haze datasets).

However, these prior-based methods usually require time-consuming iterative optimization and manually designed priors; they need to be more consistent with the practice. As is well known, haze formation is related to natural factors, such as altitude, temperature, and humidity,

**Zhiyong Tao. School of Electronic and Information Engineering, Liaoning Technical University, Huludao 125105, China.

✉ zhangshengli_win@163.com (S. Zhang); taozhiyong@lntu.edu.cn (Z. Tao); lin_sen6@126.com (S. Lin)

ORCID(s):

but it is challenging to express hazy images with a simplistic model. Therefore, these prior-based methods may result in estimation errors when dealing with complex scenes such as the non-homogeneous and dense fog weather.

Nowadays, data-driven based methods have made significant progress in image dehazing. A handful of end-to-end models [17, 6, 24, 44] have been proposed to mitigate the deep reliance on predefined prior information, but they typically have limitations in terms of interpretability. With the rapid development of deep learning techniques and the establishment of large-scale synthetic datasets [25], many data-driven approaches [50, 38, 14, 54, 46] use Convolution Neural Networks (CNNs) to extract features and build end-to-end dehazing networks to learn the transmission map between clear and hazy images.

Although the above data-driven approaches significantly improve the visual quality of image dehazing results, the following problems still exist for the current state-of-the-art (SOTA) CNNs: First, due to the complex distribution of haze in the real world, edge information is crucial to recovering the texture details of a clear image. Still, the points at the edges of the image may be ignored during the layer-by-layer convolution process, resulting in the loss of image details easily. Second, models trained on synthetic fog datasets do not work well for processing images in real-world non-homogeneous fog or dense fog, and the generalization ability and robustness of the network still need to be improved. Third, balancing the network's generalization ability with the model's complexity is a significant challenge for image dehazing tasks, especially handling challenging visual tasks.

Fig. **??** shows an example of images and their corresponding frequency information. As we observed, the low-frequency layer of the image retains more structural information, such as color and target. In contrast, the high-frequency layer of the image can specifically represent the details of the images (e.g., edges, textures). Hence, the frequency information is essential for recovering the structure and texture of an image. Therefore, we exploit the discrete wavelet transform (DWT) and inverse discrete wavelet transform (IDWT) to decompose the RGB image into high and low-frequency information to guide the network for image recovery. The motivation is that DWT or IDWT can alleviate information loss and enlarge the receptive field with a better trade-off between efficiency and restoration performance. In addition, wavelet transform has been applied to denoising tasks [57, 28, 27] using traditional methods. Utilizing wavelet transform to incorporate multi-scale information gives the network frequency analysis capabilities.

Using a combined analysis of the above, we present WaveletFormerNet to mitigate the issue in this article. We have three main goals and design three key steps to address the complex distribution of haze characteristics in the real world. Specifically: **a)** To alleviate the image's detailed texture loss due to downsampling, we devise the WaveletFormer block and IWaveletFormer block to fully preserve the structure and texture details in the original image, combining the advantages of discrete wavelet transform (DWT) and inverse discrete wavelet transform (IDWT) with the Vision Transformer (ViT), respectively. **b)** To capture the multi-frequency signals in the lightweight mechanism, we introduce parallel convolution in ViT; this structural design also reduces the computational cost and model complexity. **c)** To maintain image resolution and enhance the receptive field of our network, we present the feature aggregation module (FAM) to process the association of information and the interaction of characteristics between different levels.

Our key contributions can be summarized as follows.

1) We present WaveletFormerNet, an end-to-end Transformer-based wavelet reconstruction network to tackle image dehazing problems under complex, hazy conditions in the real world. We exploit the wavelet transform to decompose the image and allow the frequency information to guide the network in recovering the structure and texture of the image.

2) We propose the WaveletFormer and IWaveletFormer blocks to alleviate texture detail loss and maintain image resolution. The parallel convolution in the Transformer blocks captures the multi-frequency information in the lightweight mechanism, decreasing the network computational expenses.

3) We present a feature aggregation module (FAM) to capture the long-range dependencies among information with different levels and further enhance the feature extraction capability of WaveletFormerNet.

4) The satisfactory results obtained on the agricultural dust dataset (RB-Dust) [7] validate the better generalization capability of WaveletFormerNet, and the feature point matching test also exhibits that WaveletFormerNet performs better in computer vision-related applications compared to SOTA methods.

5) Extensive experimental results have demonstrated that WaveletFormerNet outperforms SOTA methods on extensive real-world fog benchmark datasets.

The remaining sections of this paper are structured as follows. Section 2 provides a concise overview of the existing research on image dehazing, wavelet transform, and ViT. In Section 3, we outline the proposed WaveletFormerNet framework, the proposed WaveletFormer and IWAveletFormer blocks, and our proposed FAM. In Section 4, we introduce the dataset details, implementation details, and the loss function we adopted. The evaluation of our proposed method is shown in Section 5, including a comparison to SOTA methods, generality analysis for WaveletFormerNet, application test, and ablation study. In Section 6, we summarize the conclusions and discussions derived from our study.

## 2. Related work

In this section, we focus on related work in image dehazing in Senction 2.1. We also introduce the related work of discrete wavelet transform in Section 2.2 and Vision Transformer in Section 2.3.

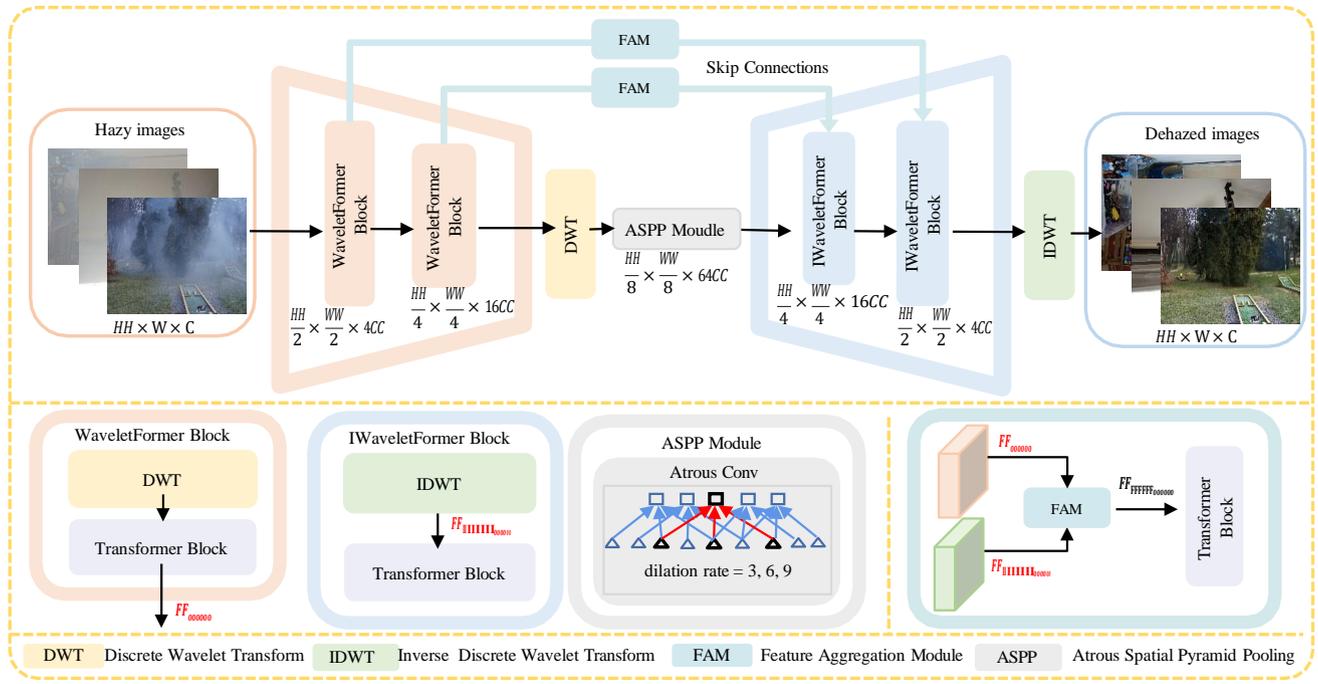

**Figure 2:** The schematic illustration of the proposed WaveletFormerNet.

## 2.1. Image dehazing

The existing methods for image dehazing are broadly classified into two categories: traditional prior-based methods and learning-based methods.

### 2.1.1. Traditional methods

Most prior-based dehazing methods [21, 17, 56, 6, 44, 55] use hazy and clear images to estimate the transmission map, then use ASM to recover haze-free images. He et al. [21] proposed the dark channel prior (DCP), assuming that the image patches of haze-free outdoor images often have low-intensity values in at least one channel. To address the difference in brightness and saturation of hazy images, Zhu et al. [56] proposed color attenuation prior (CAP) to estimate the scene depth as solid prior knowledge. To eliminate the polarization effect of information. Shen et al. [44] proposed a globally nonuniform ambient light model to predict spatially varied ambient light and designed a bright pixel index to correct the transmission. With the predicted haze parameters, they reversed the atmospheric scattering model to restore visibility. Zhou et al. [55] proposed a robust polarization-based dehazing network. However, the specific scenario inherently limits the performance of these methods, and they may lead to undesirable color distortions when the scenario does not satisfy these priors. In contrast, WaveletFormerNet can reconstruct images with richer detail by leveraging the complementary advantages of prior- and deep learning-based methods.

### 2.1.2. Deep learning methods

Recently, deep learning techniques have been proposed to tackle the problem of underwater image dehazing. These techniques have shown promising results in the restoration of underwater images. They can be classified into three categories: (i) CNN-based methods, (ii) GAN-based methods, and (iii) Transformer-based methods.

**CNN-based methods.** CNN-based methods [41, 8, 24, 29, 38, 28, 45, 50, 46, 54] have dominated in recent years. Ren et al. [41] proposed MSCNN to estimate $t(x)$ using a coarse-scale network followed by local optimization. Li et al. [24] reiterated ASM and proposed AODNet to learn each hazy image and its $t(x)$. However, all of these methods rely on ASM, and the dehazing results are often color-distorted. To alleviate the bottleneck problem encountered in traditional multi-scale methods, Liu et al. [29] implemented an attention-based end-to-end dehazing network, GridDehazeNet. To enable more efficient dehazing network performance, Liu et al. [38] designed an FFANet with channel and spatial attention to obtain excellent dehazing performance. Zheng et al. [54] took FFANet as a baseline, proposing $C^2$PNet with a curricular contrastive regularization and the physics-aware dual-branch unit to enhance the network dehazing performance. However, behind the excellent performance achieved by these supervised methods, a large number of data pairs are required for the training; more importantly, these methods are almost trained on synthetic images [25, 1, 3, 4, 2], which can not be well generalized to real-world image dehazing.

**GAN-based methods** Recently, some unsupervised data-driven methods [40, 33, 15, 18, 48, 26] have also made significant progress in image defogging. Ren et al. [40] first introduced Generative Adversarial Networks (GANs) to the field of image defogging and proposed an end-to-end defogging network that achieves mapping from foggy images to fog-free images by training a generator and a discriminator. Mehta et al. [33] propose SkyGAN for haze removal in aerial images, alleviating the degradation in image visibility. Dong et al. [15] propose a fully end-to-end GAN with a Fusion discriminator (FD-GAN) for image dehazing; this model can generate more natural and realistic dehazed images with less color distortion and fewer artifacts. Wang et al. [48] proposed a dual multiscale network, TMS-GAN, to alleviate the problem of limited domain transfer performance between trained synthetic blurred images and untrained real blurred images. Li et al. [26] propose a novel single-image dehazing algorithm by combining model-based and data-driven approaches. The proposed neural augmentation reduces the number of training data significantly, and the proposed neural augmentation framework converges faster than the corresponding data-driven approach. However, unsupervised methods may be unstable during the training process, leading to problems such as the possibility of unstable results during enhancements. .
**Transformer-based methods** Recently, Transformer [47] has gained increasing attention, image content and attention weights interact spatially as a result of spatially varying convolution. Guo et al. [19] proposed DeHamer to effectively integrate Transformer features and CNN features and bring the domain knowledge, such as task-specific prior, into Transformer for improving the performance. Furthermore, Song et al. [46] proposed that DehazeFormer improves on Swin Transformer [30], which makes Transformer more useful for image dehazing. Compared to CNN-based networks, our approach can help the network pay more attention to attenuated color channels and spatial areas. In addition, since a GAN is coupled with a transformer, we can obtain a better performance with a relatively small number of parameters.

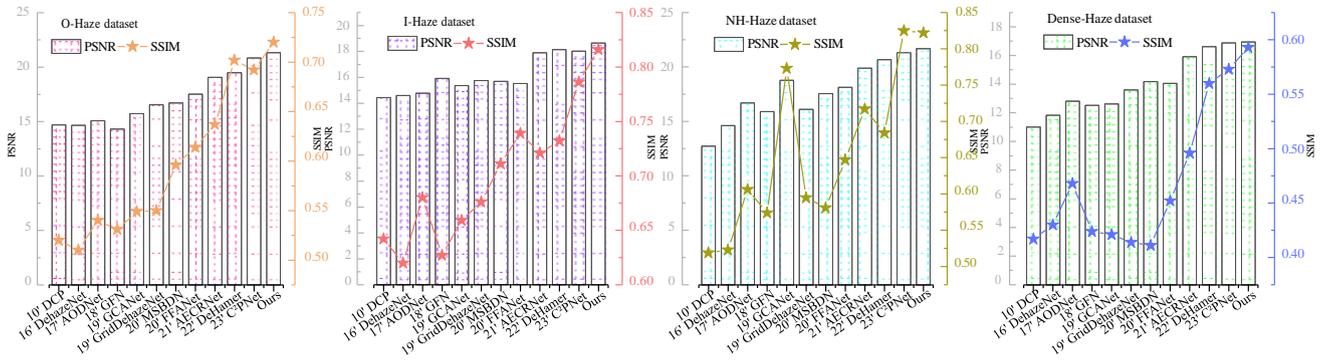

**Figure 3:** Quantitative comparisons on referenced indicators (PSNR and SSIM) results among WaveletFormerNet and SOTA methods on four real-world datasets ((O-Haze, I-Haze, NH-Haze, and Dense-Haze)).

### 2.2. Discrete wavelet transform

As a traditional image processing technique, the discrete wavelet transform [12, 52, 13, 28] is widely used for image analysis. Guo et al. [20] proposed a DWSR combining the discrete wavelet transform with ResNet by predicting the residual wavelet subbands. Inspired by U-Net [43], Liu et al. [27] proposed MWCNN, which replaces pooling and non-pooling operations to reduce the number of parameters in the network. However, multiple uses of the discrete wavelet transform operations may result in redundant channels. Therefore, Yang et al. [51] proposed the Wavelet U-Net, which uses the discrete wavelet transform to extract edge features while applying the adaptive color transform that convolutional layers; this structure enhances the texture details in the image. Zou et al. [57] proposed SDWNet to obtain large sensory fields with a high spatial resolution and recover precise high-frequency texture details. Fu et al. [18] proposed a two-branch network DW-GAN to leverage the power of discrete wavelet transform in helping the network acquire more frequency domain information. These methods demonstrate the significant role of discrete wavelet transform in the image recovery process.

### 2.3. Vision Transformer

Attention mechanism [47] of deep learning has achieved a great process nowadays. Dosovitskiy et al. [16] proposed ViT with the direct application of the Transformer architecture, which projects images into token sequences via patch-wise linear embedding. The shortcomings of the ViT are its weak inductive bias and its quadratic computational cost. Until Liu et al. [30] proposed the Swin Transformer, they divided tokens into a window and performed self-attention within a window to maintain a linear computational cost. Guo et al. [19] proposed Dehamer modulate convolutional features via learning modulation matrices, which are conditioned on Transformer features instead of simple addition or concatenation of features. Song et al. [46] proposed the DehazeFormer, which can be viewed as a combination of Swin Transformer [30] and U-Net [43] with more comprehensive improvements in the normalization layer, nonlinear activation function, and spatial information aggregation scheme. DehazeFormer improves the network performance for single-image dehazing further. Although ViT enhances the image recovery performance, it may increase additional computational expense and ignore the haze distribution characteristics under complex conditions.

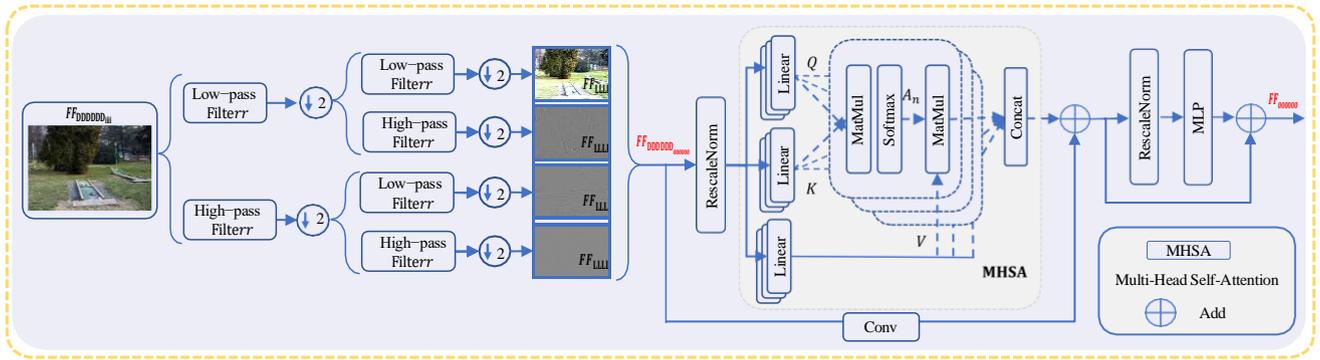

**Figure 4:** The architecture of proposed WaveletFormer and IWaveletFormer blocks. Note: The WaveletFormer block and the IWaveletFormer block have the same structure: they utilize DWT and IDWT to substitute downsampling and upsampling, respectively.

## 3. Methodology

In this section, We introduced the motivation and overview of our proposed WaveletFormerNet in Section 3.1; in SectionSection 3.2, we present the detailed structure of the WaveletFormer and IWaveletFormer blocks. The detailed design of FAM, another module proposed in this paper, is placed in Section 3.3. At the end of the chapter, we introduce the ASPP Module we adopted in Section 3.4.

### 3.1. Overview

Fig. 2 illustrates the detailed structure of our WaveletFormerNet. Both encoding and decoding of WaveletFormerNet are based on the WaveletFormer and IWaveletFormer block, but the difference between the encoding and decoding segments is that downsampling and upsampling are replaced by DWT and IDWT, respectively. Although the WaveletFormer and IWaveletFormer block as the base block of the network mainly combines wavelet transform and Swin Transformer, we do not directly apply these two existing tools but improve them. We use the wavelet transform to transform the features to the frequency domain and use the frequency information to guide WaveletFormerNet to recover the structural and texture details of the image. In addition, our proposed parallel convolution also alleviates the receptive field caused by Swin Transformer. This structure of the proposed WaveletFormer and IWaveletFormer block also alleviates the details caused by downsampling loss and other problems. Furthermore, we propose a Feature Aggregation Module (FAM) to maintain image resolution and enhance the receptive field of our network, combining different levels of feature information. Finally, we adapt an atrous spatial pyramid pooling module (ASPP) in our network and adjust dilated convolution [10] with different expansion rates (rate = 3, 6, 9), obtaining features in different receptive fields.

### 3.2. WaveletFormer and IWaveletFormer block

WaveletFormer and IWaveletFormer Blocks use DWT and IDWT to decompose the images from the frequency domain point of view, respectively, and the feature maps are used as inputs to the Transformer module with parallel convolution.

#### 3.2.1. Frequency decomposition of images.

Fig. 4 illustrates the detailed structure of the WaveletFormer block, adopting frequency information to guide the network in reconstructing a clear image. We can observe that the input image can be divided into the low- and high-frequency details separated into four different frequency subbands: the low-frequency band the horizontal subband $F_{LH}$, the vertical subband and the high-frequency subband on the diagonal edge of the original image. This mechanism alleviates detail and color loss and provides a better balance between network processing efficiency and image recovery performance. For the 2D discrete wavelet transform, we import the pytorch wavelets package and use Daubechies wavelet basis functions.

#### 3.2.2. Parallel convolution in Vision Transformer.

According to the attention mechanism [47], Fig. 4 illustrates the detailed structure of our WaveletFormer block.

We perform an additional convolution of the features from the DWT and then achieve a dynamic aggregation style of information with the production of MHSA in the spatial dimension, thus capturing the multi-frequency signals in the lightweight mechanism. Therefore, the output of the WaveletFormer block can be formulated as follows:

In the decoding, we use the IDWT in the IWaveletFormer block with the same filter as the DWT for image recovery.

### 3.3. Feature aggregation module

Fig. 5 illustrates the detailed structure of the proposed FAM. As the key component of the FAM, the Multi-Head Cross-Attention (MHCA) [37] introduces the feature map of the WaveletFormer block and the feature map from IWaveletFormer block into the MHSA [47] for processing, the computed weight values $Y$ to be rescaled by the sigmoid activation function. The resulting feature tensor $Z$ will be summed with the feature map $F_{out}$ to obtain the high-level feature tensor.

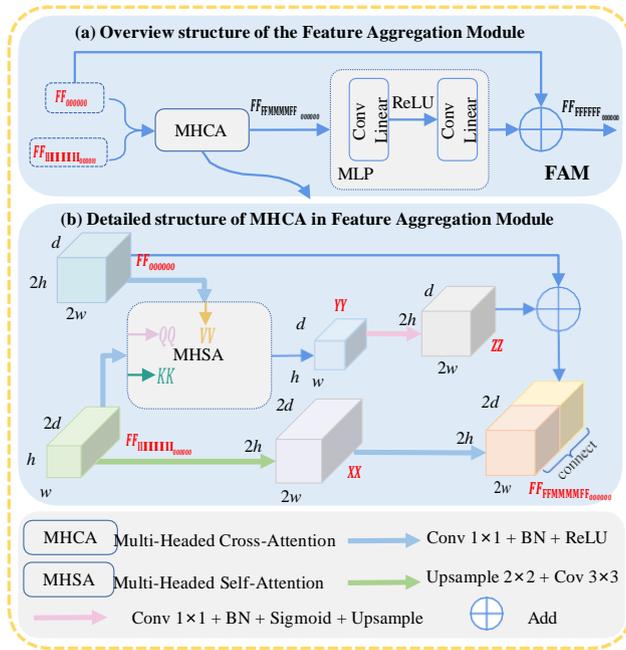

**Figure 5:** Architecture of proposed feature aggregation module employed in the proposed WaveletFormerNet.

The proposed feature aggregation module (FAM) introduces the feature map of the WaveletFormer block and the feature map from IWaveletFormer block for processing. Fig.6 illustrates that FAM is a link between the encoding and decoding stages, guiding our WaveletFormerNet to generate images with more crisp textures and rich details. The proposed FAM removes irrelevant or noisy areas from skip connections and highlights those important areas, capturing the long-range relationship among different receptive field features and improving decoding efficiency.

### 3.4. Atrous Spatial Pyramid Pooling Module

We adopt the atrous spatial pyramid pooling module (ASPP module) in our network. Unlike previous dehazing networks that use repeated upsampling and downsampling to obtain large receptive domains, we use dilated convolution to obtain features in different receptive fields. Fig. 7 illustrates the principle of the ASPP module used in our paper. The input is sampled in parallel with a convolution of holes at different expansion rates (rate = 3, 6, 9); the results are then concatenated together to expand the number of channels, and the channels of the output are reduced by 1 × 1 convolution.

## 4. Experiment setup

This section introduces the training and test datasets we use in Section 4.1 and our datasets expansion way in Section 4.1.1. Section **??** introduces our loss function and training details in Section 4.2. Finally, our comparison methods and evaluation metrics are embodied in Section 4.3.

### 4.1. Training datasets

We extensively and comprehensively evaluated our model and compared SOTA methods on real-world and synthetic datasets in the same experiment setting.

#### 4.1.1. Real-world datasets

We use the following four datasets to evaluate our experiments: the NTIRE 2018 image dehazing dataset (I-Haze), the outdoor NTIRE 2018 image dehazing dataset (O-Haze), a benchmark for image dehazing with dense-haze and haze-free images (Dense-Haze), and the NTIRE 2020 dataset for non-homogeneous dehazing challenge (NH-Haze).

**I-Haze [1] and O-Haze [3]:** They contain 25 and 35 hazy images (size 2833 × 4657 pixels) respectively for training. Both datasets contain 5 hazy images for validation along with their corresponding ground truth images. We used training data for training and validation data for the test.

**Dense-Haze [2]:** It contains 45 hazy images (size 1200 × 1600 pixels) for training 5 hazy images for validation and 5 more for testing with their corresponding ground truth images. We have performed training on training data and tested our model with test data.

**NH-Haze [4]:** It contains 45 hazy images (size 1200 × 1600 pixels) for training. We selected 40 pairs of data for training and the rest for testing.

**Real-world Datasets Expansion**

We use the same training strategy and dataset expansion process for all four real-world datasets. Specifically, we randomly cropped the original images into square patches of 512 × 512 pixels; these patches are not the same for every epoch. To augment the training data, we implemented random rotations (90, 180, or 270 degrees) and random horizontal flips when processing the training data. This step allows these small real-world datasets to be expanded into larger datasets that are more efficient and more suitable for data-driven methods of training experiments.

#### 4.1.2. Synthetic datasets

For the objectivity of the experimental results, we evaluated our and SOTA methods under the same experimental conditions. We train our WaveletFormerNet on two training sets of RESIDE: indoor and outdoor. The indoor training set (ITS) has 13,990 hazy images, and the outdoor training set (OTS) has 296,695 hazy images. We selected SOTS as our testing set, the SOTS is from the RESIDE dataset, including 500 indoor and 500 outdoor hazy images.

### 4.2. Loss function and training details

Referring to previous work [53], to balance both visual perception and quantitative assessments, we combine $\ell_1$ loss, multiscale structural similarity (MS-SSIM) loss, and perceptual loss function linearly.

Concretely, the $\ell_1$ loss retains color and brightness and converges quickly, providing a wider and more stable gradient. The MS-SSIM loss integrates the variations of resolution and visualization conditions to consider structural differences, compared to other loss functions, preserving the contrast in the high-frequency region.

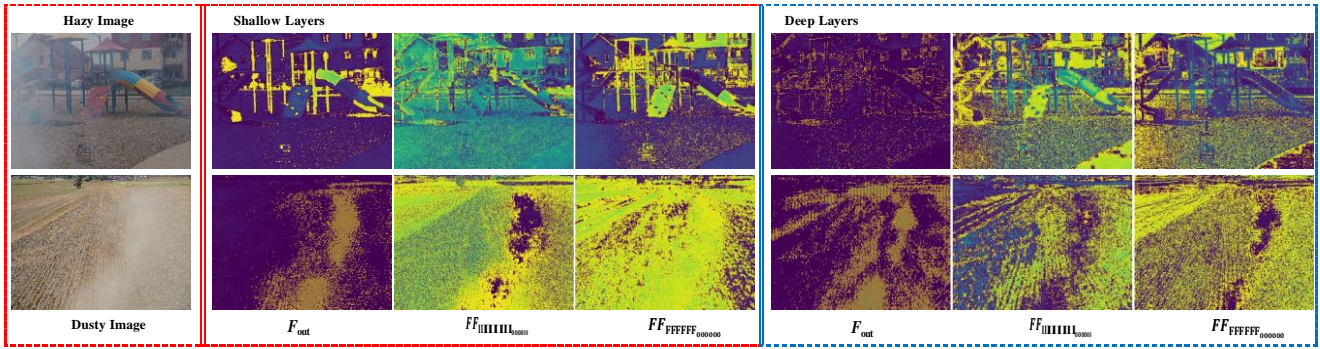

**Figure 6:** Visual results of the intermediate features in our proposed feature aggregation module with degraded images (hazy and dusty images). The corresponding modulated features $F_{\text{FAM}_{\text{out}}}$ are also presented. The features $F_{\text{out}}$ in the WaveletFormer block have long-range attention but coarse textures, while the features $F_{\text{IDWT}_{\text{out}}}$ in the IWaveletFormer block are with precise details. The modulated features produced by the feature aggregation module inherit the characteristics of both Transformer features and frequency information, i.e., long-range dependencies and clear textures.

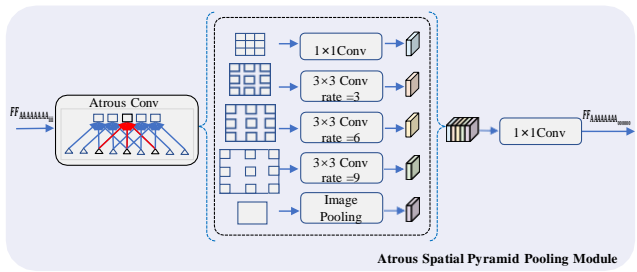

**Figure 7:** The architecture of our adopted Atrous Spatial Pyramid Pooling Module (ASPP Module).

Inspired by the current hot research in the field of image dehazing, we adopt perceptual loss to promote the perceptual similarity of dimensional spatial features and perceive the image from a high dimension.

We conducted a comparative experiment on GeForce RTX 4090 using PyTorch 1.11.0. Adam optimizer is adopted, the initial learning rate is set to 0.0001, betas = (0.9, 0.999), the batch size is 16, the crop size is 256 × 256, and the total number of epochs is 150. We adopt the cosine annealing strategy [22] to adjust the learning rate $\eta_t$ from the initial value to 0.

### 4.3. Comparison methods and evaluation metric

We conduct a comprehensive comparison with SOTA methods on synthetic and real-world datasets, and we mainly select some SOTA methods for comparison according to the three main categories of image dehazing methods.

We adopt two objective quantity evaluation metrics ( peak signal-to-noise ratio (PSNR) and structural similarity index (SSIM)). In addition, we also adopt three no-reference metrics ( Entropy [5], natural image quality evaluator ( NIQE [34]), and fog aware density evaluator ( FADE [11])) to evaluate all the above methods. Specifically, a higher Entropy score indicates that the image presents more detail; a lower NIQE score indicates better image quality and a lower FADE score indicates better visibility.

## 5. Experiment results

We show the results of our proposed WaveletFormerNet compared to the SOTA method on real-world and synthetic datasets in Section 5.1 and Section 5.2, respectively. In Section 5.3, we show the parameters and runtime analysis. We display a generality analysis for WaveletFormerNet on the RB-Dust agricultural dust dataset in Section 5.4. We also conduct the application test in Section 5.5 and the ablation study in Section 5.6.

### 5.1. Results analysis on the real-world datasets

We show the comparative results of qualitative effects on the four datasets in Fig. 9, Fig. 8, Fig. 10, and Fig. 11. We can observe that DCP produces bluer results in real-world datasets, with severe color bias due to the complex hazy conditions where the pure prior theory is not applicable. Furthermore, the output results of AODNet, FFANet, and AECRNet show severe color distortion and incomplete haze removal; Dehamer and $C^2$PNet outperform the above three methods, removing haze very well and providing enjoyable visual effects. However, our results are closer to the ground truth than these two SOTA methods, and WaveletFormerNet produces visually pleasing dehazing images, which can retain richer texture details.

Table 1 shows the quantitative comparison of WaveletFormerNet with SOTA methods on referenced indicators (PSNR and SSIM). We can observe that methods such as AECRNet, Dehamer, and $C^2$PNet have demonstrated exemplary performance. However, for the challenging tasks of non-homogeneous fog and dense fog, our WaveletFormerNet improves the PSNR metrics by 0.36dB and 0.07dB each compared to the second-best on the NH-Haze and Dense-Haze datasets.

Moreover, Table 2 shows the quantitative comparison of WaveletFormerNet with SOTA methods on non-referenced indicators (Entropy, NIQE, and FADE). The lowest NIQE and FADE metrics and the highest Entropy metric illustrate the better image quality and visibility produced by our WaveletFormerNet.

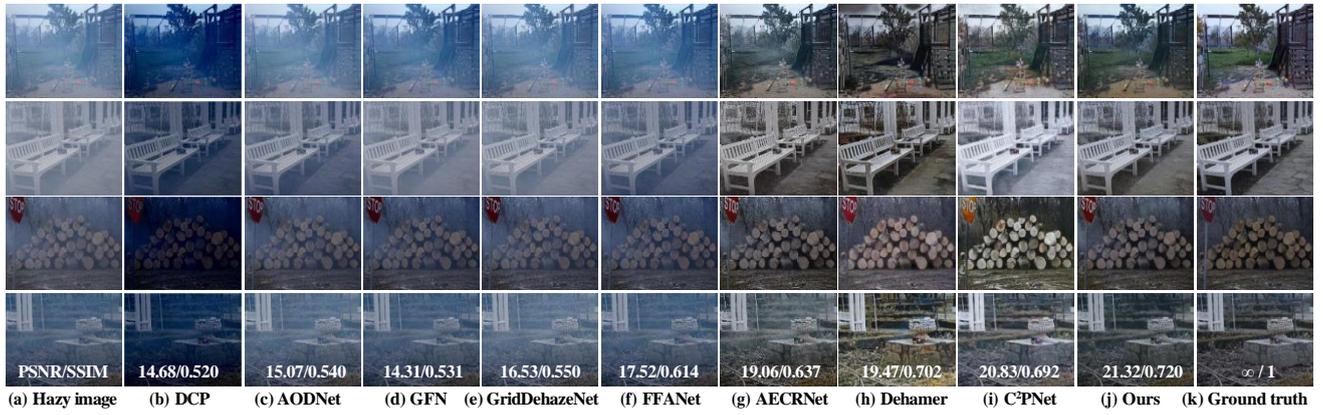

**Figure 8:** Qualitative comparison results among WaveletFormerNet and SOTA methods on the real-world fog O-Haze dataset.

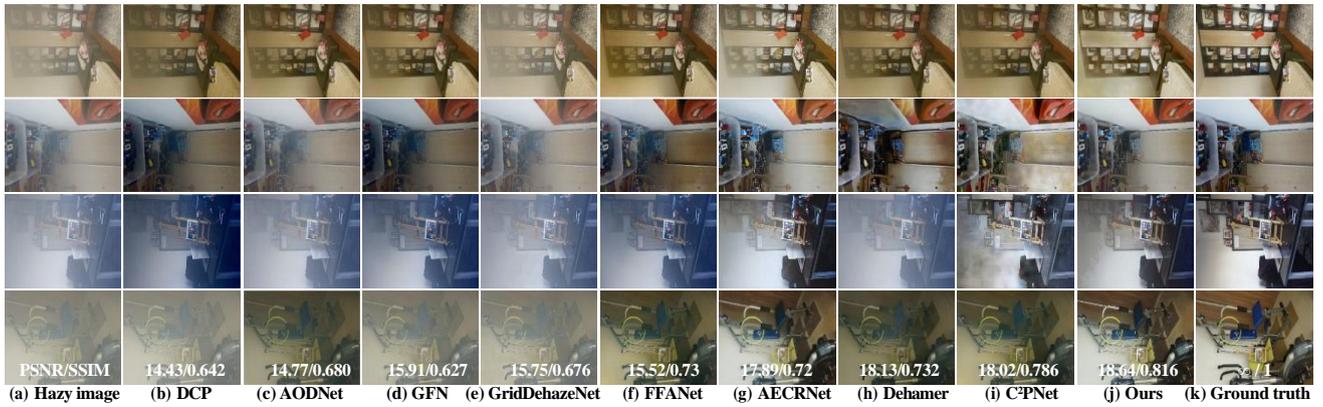

**Figure 9:** Qualitative comparison results among WaveletFormerNet and SOTA methods on the real-world fog I-Haze dataset.

## 5.2. Results analysis on the synthetic dataset.

Table 3 and Fig. 12 show that our proposed WaveletFormerNet produces competitive comparison results with SOTA methods on SOTS datasets. Compared to FFANet, DCP and AODNet do not remove fog very thoroughly; Dehamer and DehazeFormer-B are capable of outputting higher quality images; $C^2$PNet leads all the methods in terms of objective evaluation metrics, but our method produces competitive comparisons with fewer parameters than $C^2$PNet on the qualitative aspects while maintaining richer details and color information.

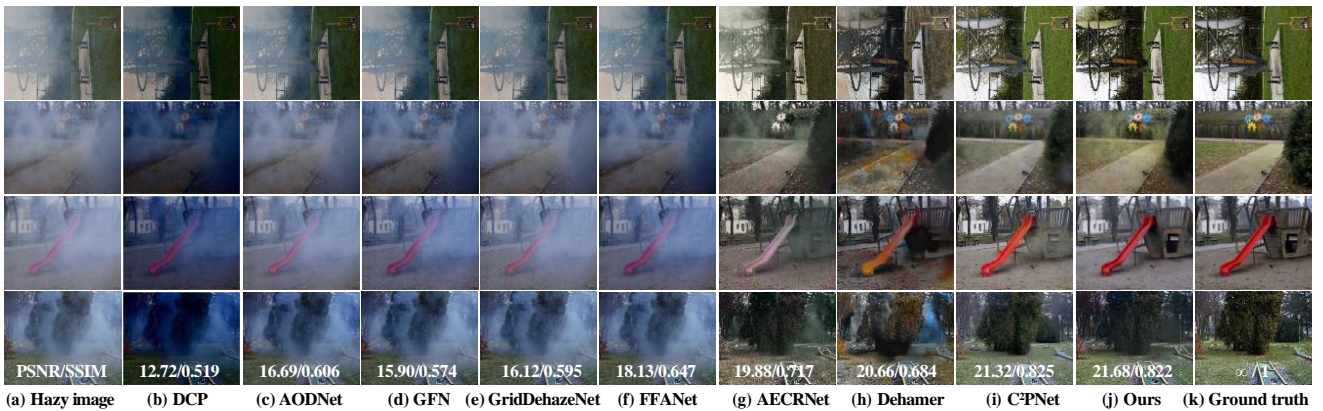

**Figure 10:** Qualitative comparison results among WaveletFormerNet and SOTA methods on the real non-homogeneous fog NH-Haze dataset.

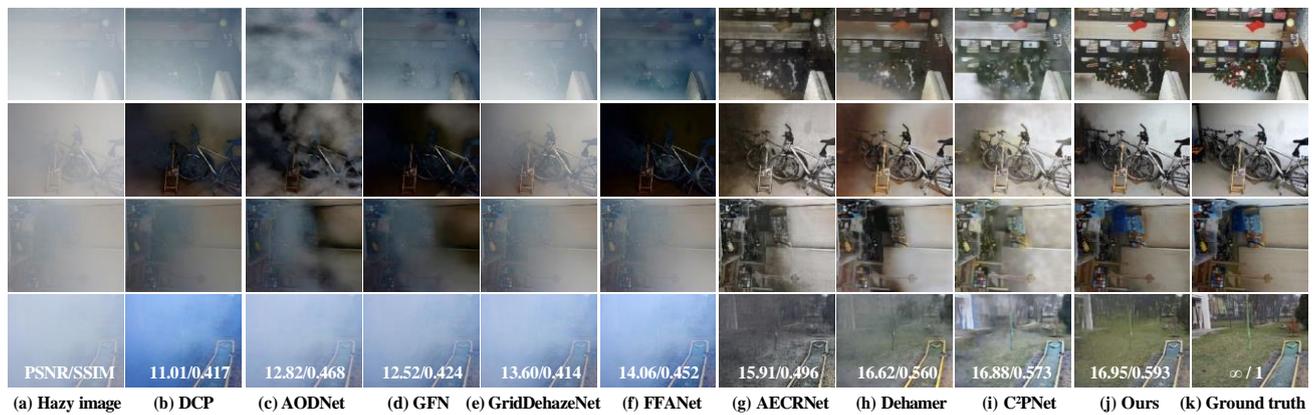

**Figure 11:** Qualitative comparison results among WaveletFormerNet and SOTA methods on the real-world dense fog Dense-Haze dataset.

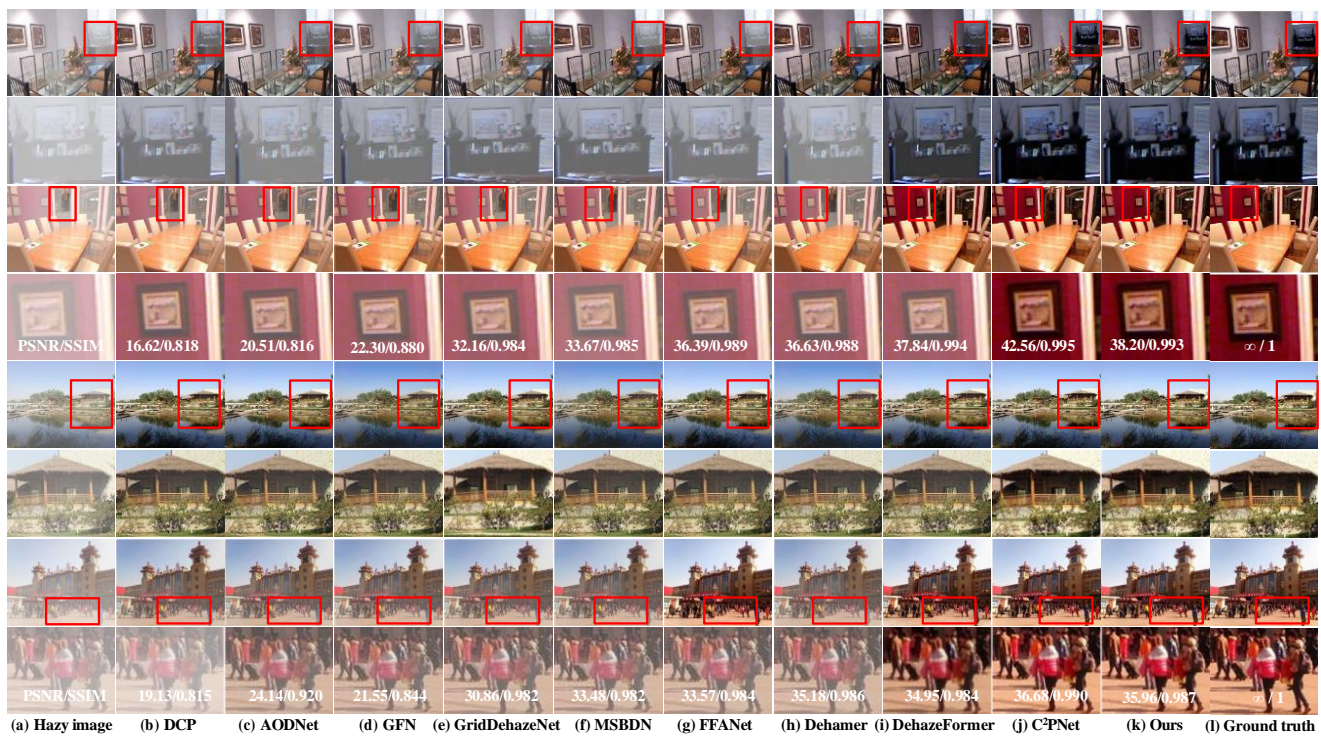

**Figure 12:** Qualitative comparison results among WaveletFormerNet and SOTA methods on the synthetic hazy datasets (SOTS-indoor and SOTS-outdoor). The red frame lines represent enlarged details from the original images.

Therefore, combining the quantitative and qualitative results from the real-world and synthetic datasets, our WaveletFormerNet provides a more complete balance between dehazing performance and model complexity.

### 5.3. Parameters and runtime analysis

Table 4 demonstrates that our method has a significant advantage over the SOTA methods in parameters because decomposing the image by wavelet transform in high and low frequency before processing reduces the network overhead. However, our approach is not superior in inference time because the wavelet transform takes up some time.

However, WaveletFormerNet outperforms the SOTA methods on real-world datasets, which explains a comprehensive view of model complexity and dehazing performance.

### 5.4. Generality analysis for WaveletFormerNet

To our knowledge, the RB-Dust dataset [7] is the first publicly available agricultural landscape dusting dataset, consisting of 200 images with 1920 × 1080 pixels. Previous work by Peter et al. [7] demonstrated that dust properties are similar to haze properties and that some image dehazing algorithms are also suitable for image dedusting. Therefore, we selected the classical algorithms for image defogging in recent years to compare the generalizability among WaveletFormerNet and other SOTA methods. Fig. 14 shows more qualitative comparison results among WaveletFormerNet and SOTA methods, which illustrates that WaveletFormerNet removes more dense and non-homogeneous dust and retains more textural detail, demonstrating the promising robustness and better generalization ability of the proposed WaveletFormerNet.

**Table 1**
Quantitative comparisons on referenced indicators between WaveletFormerNet and SOTA methods on the real-world datasets (I-Haze, O-Haze, Dense-Haze, and NH-Haze). Indicators marked with ↑ indicate higher and better data; ↓ indicate lower and better. We use **bold** and underline to mark the best and second-best methods.

| Methods | O-Haze | | I-Haze | | Dense-Haze | | NH-Haze | |
|---|---|---|---|---|---|---|---|---|
| | PSNR↑ | SSIM↑ | PSNR↑ | SSIM↑ | PSNR↑ | SSIM↑ | PSNR↑ | SSIM↑ |
| TAPAMI'10 DCP [21] | 14.68 | 0.520 | 14.43 | 0.642 | 11.01 | 0.417 | 12.72 | 0.519 |
| TIP'16 DehazeNet [8] | 14.65 | 0.510 | 14.59 | 0.620 | 11.84 | 0.430 | 14.62 | 0.523 |
| ICCV'17 AODNet [24] | 15.07 | 0.540 | 14.77 | 0.680 | 12.82 | 0.468 | 16.69 | 0.606 |
| CVPR'18 GFN [42] | 14.31 | 0.531 | 15.91 | 0.627 | 12.52 | 0.424 | 15.90 | 0.574 |
| WACV'19 GCANet [9] | 15.71 | 0.549 | 15.37 | 0.659 | 12.62 | 0.421 | 18.79 | 0.773 |
| ICCV'19 GridDehazeNet [29] | 16.53 | 0.550 | 15.75 | 0.676 | 13.60 | 0.414 | 16.12 | 0.595 |
| CVPR'20 MSBDN [14] | 16.69 | 0.596 | 15.68 | 0.711 | 14.18 | 0.411 | 17.54 | 0.581 |
| AAAI'20 FFANet [38] | 17.52 | 0.614 | 15.52 | 0.739 | 14.06 | 0.452 | 18.13 | 0.647 |
| CVPR'21 AECRNet [50] | 19.06 | 0.637 | 17.89 | 0.721 | 15.91 | 0.496 | 19.88 | 0.717 |
| CVPR'22 DeHamer [19] | 19.47 | <u>0.702</u> | <u>18.13</u> | 0.732 | 16.62 | 0.560 | 20.66 | 0.684 |
| CVPR'23 C$^2$PNet [54] | <u>20.83</u> | 0.692 | 18.02 | <u>0.786</u> | <u>16.88</u> | <u>0.573</u> | <u>21.32</u> | **0.825** |
| **WaveletFormerNet(Ours)** | **21.32** | **0.720** | **18.64** | **0.816** | **16.95** | **0.593** | **21.68** | <u>0.822</u> |

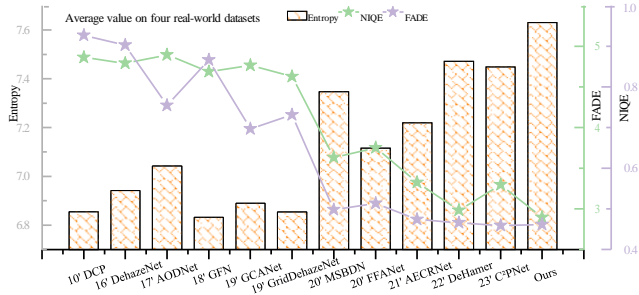

**Figure 13:** Quantitative comparisons on non-referenced indicators (Entropy, NIQE and FADE) results among WaveletFormerNet and SOTA methods on four real-world datasets ((O-Haze, I-Haze, NH-Haze, and Dense-Haze)). The data in the table represent the average of the three indicators over all real-world datasets.

**Table 2**
Quantitative comparisons on non-referenced indicators between WaveletFormerNet and SOTA methods on the real-world datasets (I-Haze, O-Haze, Dense-Haze, and NH-Haze). Indicators marked with ↑ indicate higher and better data; ↓ indicate lower and better. We use **bold** and underline to mark the best and second-best methods, where the no-reference metrics (Entropy, NIQE, FADE) are averaged over four real-world datasets.

| Methods | Entropy↑ | NIQE↓ | FADE↓ |
|---|---|---|---|
| TAPAMI'10 DCP [21] | 6.855 | 4.866 | 0.927 |
| TIP'16 DehazeNet [8] | 6.942 | 4.792 | 0.903 |
| ICCV'17 AODNet [24] | 7.043 | 4.897 | 0.755 |
| CVPR'18 GFN [42] | 6.832 | 4.689 | 0.867 |
| WACV'19 GCANet [9] | 6.890 | 4.766 | 0.697 |
| ICCV'19 GridDehazeNet [29] | 6.854 | 4.627 | 0.731 |
| CVPR'20 MSBDN [14] | 7.347 | 3.631 | 0.498 |
| AAAI'20 FFANet [38] | 7.115 | 3.754 | 0.514 |
| CVPR'21 AECRNet [50] | 7.220 | 3.328 | 0.474 |
| CVPR'22 DeHamer [19] | <u>7.472</u> | <u>2.983</u> | 0.466 |
| CVPR'23 C$^2$PNet [54] | 7.449 | 3.297 | **0.459** |
| **WaveletFormerNet(Ours)** | **7.631** | **2.897** | <u>0.462</u> |

### 5.5. Application test

The SIFT algorithm [31] is used to detect and describe the matching of feature points between different images by extracting the local features of the image. This approach has a wide range of applications in the fields of target recognition and target tracking. In this section, we perform a feature point matching test to evaluate the performance of WaveletFormerNet. We selected the last four years of SOTA methods for comparison. Fig. 15 shows more of the application test results among our method and SOTA methods, and we can observe that WaveletFormerNet has the highest number of matching points. The application test shows that WaveletFormerNet exhibits better performance in computer vision-related applications.

### 5.6. Ablation study

To verify the effectiveness of the proposed WaveletFormerNet, we performed a structure and loss function ablation study on the NH-Haze dataset. Here, we need to emphasize the following:

1) w/o DWT and IDWT represent WaveletFormerNet without the wavelet transform and inverse wavelet transform for upsampling and downsampling.

**Table 3**

Quantitative comparisons between WaveletFormerNet and SOTA methods on the synthetic dataset (RESIDE). Indicators marked with ↑ indicate higher and better data; ↓ indicate lower and better. We use **bold** and underline to mark the best and second-best methods. Data marked with - is unavailable, where the no-reference metrics (Entropy, NIQE, FADE) are averaged over SOTS-indoor and SOTS-outdoor datasets.

| Methods | SOTS-Indoor | | SOTS-Outdoor | | Entropy↑ | NIQE↓ | FADE↓ |
|---|---|---|---|---|---|---|---|
| | PSNR↑ | SSIM↑ | PSNR↑ | SSIM↑ | | | |
| TAPAMI'10 DCP [21] | 16.62 | 0.818 | 19.13 | 0.815 | 6.349 | 4.912 | 0.421 |
| TIP'16 DehazeNet [8] | 19.82 | 0.821 | 24.75 | 0.927 | 6.570 | 4.868 | 0.407 |
| ICCV'17 AODNet [24] | 20.51 | 0.816 | 24.14 | 0.920 | 6.819 | 4.210 | 0.381 |
| CVPR'18 GFN [42] | 22.30 | 0.880 | 21.55 | 0.844 | 6.372 | 4.596 | 0.326 |
| WACV'19 GCANet [9] | 30.06 | 0.960 | 22.76 | 0.889 | 6.437 | 4.596 | 0.326 |
| ICCV'19 GridDehazeNet [29] | 32.16 | 0.984 | 30.86 | 0.982 | 6.919 | 4.429 | 0.358 |
| CVPR'20 MSBDN [14] | 33.67 | 0.985 | 33.48 | 0.982 | 6.536 | 4.013 | 0.317 |
| AAAI'20 FFANet [38] | 36.39 | 0.989 | 33.57 | 0.984 | 6.817 | 3.898 | 0.294 |
| CVPR'21 AECRNet [50] | 37.17 | 0.990 | - | - | 7.290 | 3.978 | 0.261 |
| CVPR'22 DeHamer [19] | 36.63 | 0.988 | 35.18 | 0.986 | 7.561 | **3.406** | 0.280 |
| TIP'23 Dehazeformer-B [46] | 37.84 | 0.994 | 34.95 | 0.989 | 7.310 | 3.591 | 0.253 |
| CVPR'23 C²PNet [54] | **42.56** | **0.995** | **36.68** | **0.990** | 7.603 | 3.670 | **0.249** |
| **WaveletFormerNet(Ours)** | 38.20 | 0.993 | 35.96 | 0.987 | **7.816** | 3.429 | 0.295 |

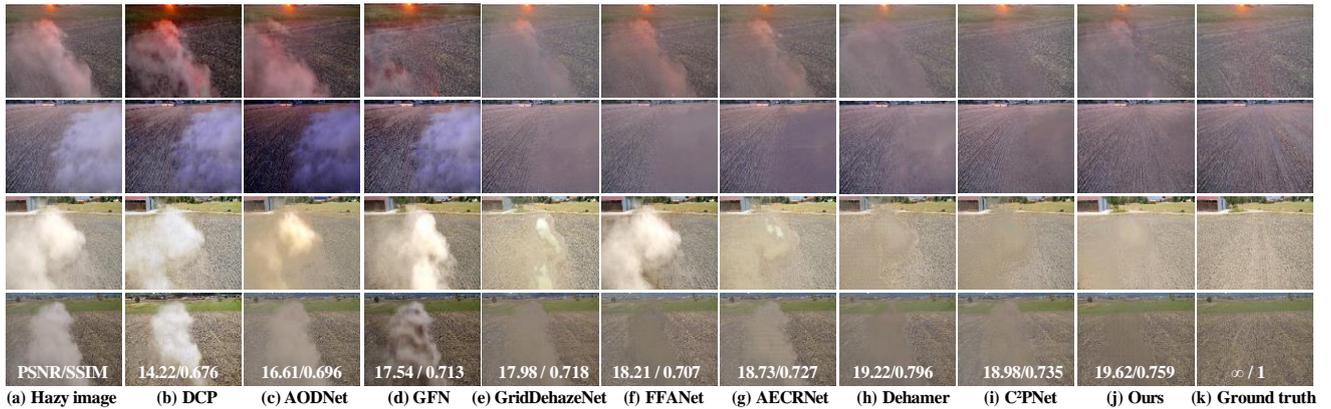

(a) Hazy image (b) DCP (c) AODNet (d) GFN (e) GridDehazeNet (f) FFANet (g) AECRNet (h) Dehamer (i) C²PNet (j) Ours (k) Ground truth

PSNR/SSIM: 14.22/0.676, 16.61/0.696, 17.54/0.713, 17.98/0.718, 18.21/0.707, 18.73/0.727, 19.22/0.796, 18.98/0.735, 19.62/0.759, ∞/1

**Figure 14:** Qualitative comparison results among WaveletFormerNet and SOTA methods on real-world dust dataset (RB-Dust dataset). The red frame lines represent enlarged details from the original images.

2) w/o parallel convolution represents the proposed WaveletFormer and IWaveletFormer block without the parallel convolution in the Transformer block.

3) w/o FAM represents WaveletFormerNet without the feature aggregation module.

4) w/o ASPP represents WaveletFormerNet without the ASPP Module.

5) the full model represents our complete WaveletFormerNet.

Table 5 and Fig. 16 show the results of the ablation study; we can observe that using the wavelet transform instead of upsampling and downsampling can be more effective in extracting texture details and the overall structure of the image from the complex haze background; in addition, our proposed FAM restores the color and brightness of the overall image, which plays a significant role in bringing the processing results closer to the ground truth; both our proposed parallel convolution and the used ASPP Module enhance the overall dehazing performance of the network.

Moreover, the results of the ablation experiments regarding the loss function are placed in Table 5; the combination of loss functions also brings significant gains in the quantitative evaluation metrics of the network.

## 6. Conclusion and discussion

The paper proposes WaveletFormerNet for real-world single-image dehazing, trading off the dehazing performance, generalization ability, and model complexity. We embed the wavelet transform into the ViT by presenting the WaveletFormer and IWaveletFormer blocks, alleviating structure and texture detail loss during the encoder and decoder. We devise FAM to capture the long-range dependencies among different levels of information and improve decoder efficiency. Extensive experiments demonstrate that our WaveletFormerNet outperforms SOTA methods on real-world fog datasets. Moreover, generality analysis and application tests show that WaveletFormerNet exhibits better generalization capability and superior performance in computer vision-related applications of our method.

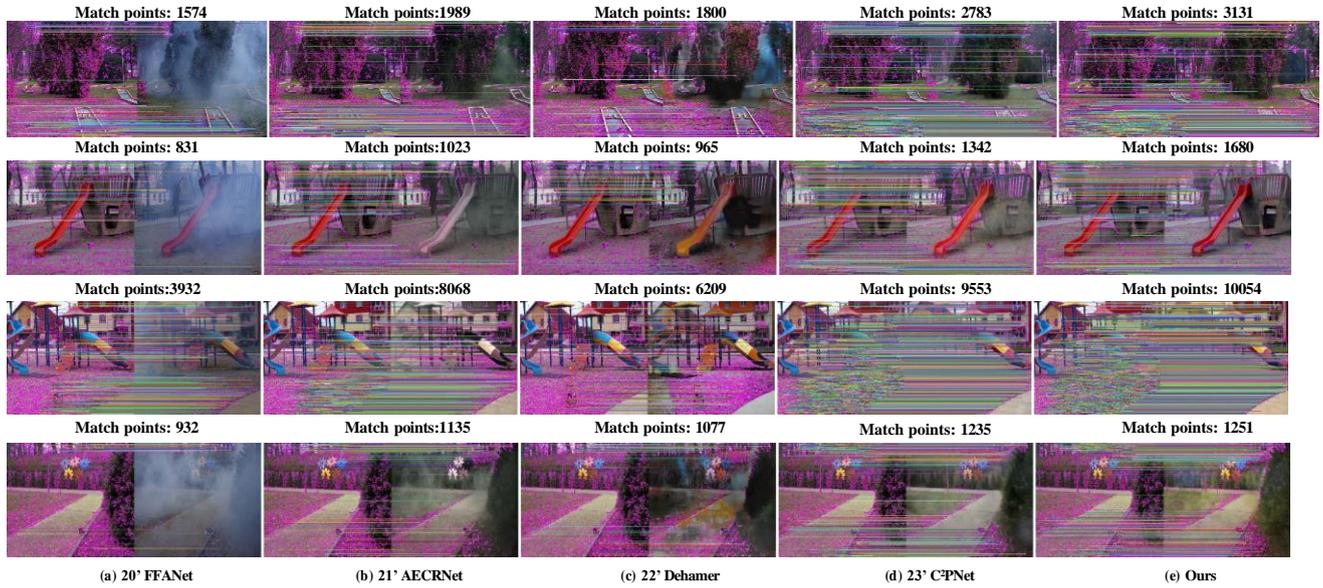

**Figure 15:** Application test results of WaveletFormerNet and SOTA methods on NH-Haze dataset. The purple dots represent feature points, and the horizontal lines represent the matching of feature points between the dehazed result by different methods (right one) and a clear reference image (left one); the denser the matching lines are, the higher the degree of feature matching.

**Table 4**
We conduct parameters (# Param), floating-point operations (# FLOPs), and inference time as the main metrics of computational efficiency on RGB image with a resolution of 256 × 256 between WaveletFormerNet and SOTA methods.

| Methods | Overhead | | |
|---|---|---|---|
| | #Param↓ | #FLOPs | Runtime↓ |
| ICCV'17 AODNet | **0.002M** | 0.115G | **0.316ms** |
| ICCV'19 GridDehazeNet | 0.956M | 21.49G | 15.35ms |
| CVPR'20 MSBDN | 31.35M | 24.44G | 9.826ms |
| AAAI'20 FFANet | 4.456M | 287.5G | 52.76ms |
| CVPR'21 AECRNet | 2.611M | 52.20G | - |
| CVPR'22 DeHamer | 132.45M | 48.93G | 26.31ms |
| TIP'23 DehazeFormer-B | 2.514M | 25.79G | 19.22ms |
| CVPR'23 C$^2$PNet | 7.17M | 429.52G | - |
| **WaveletFormerNet (Ours)** | 2.26M | 4.08G | 16.58ms |

**Table 5**
Structure and loss function ablation study of WaveletFormerNet on the NH-Haze Dataset, "w/o" means without.

| Baseline | PSNR↑ | SSIM↑ | NH-Haze dataset Loss Function | | | |
|---|---|---|---|---|---|---|
| w/o DWT and IDWT | 18.06 | 0.687 | $\ell_1$ loss | ✓ | ✓ | ✓ |
| w/o parallel convolution | 18.39 | 0.710 | MS-SSIM loss | | ✓ | ✓ |
| w/o FAM | 20.22 | 0.788 | perceptual loss | | | ✓ |
| w/o ASPP Module | 21.17 | 0.809 | PSNR | 19.60 | 21.49 | 21.68 |
| **full model (ours)** | **21.68** | **0.822** | SSIM | 0.783 | 0.817 | 0.822 |

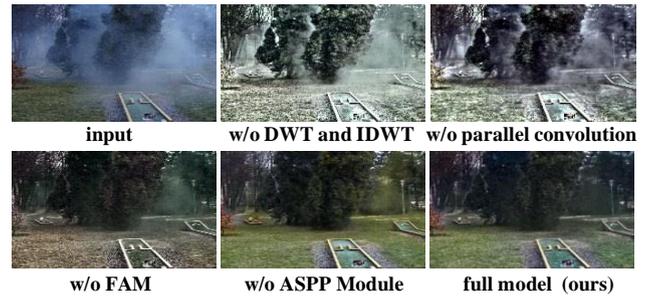

**Figure 16:** Visualisation comparison of different variants of WaveletFormerNet on the NH-Haze dataset.

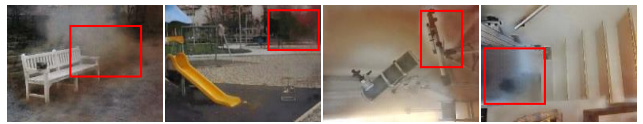

**Figure 17:** Some output results of WaveletFormerNet on the Dense-Haze dataset may result in artifacts or incomplete defogging. The red frame represents a zoomed-in detail.

Although the proposed WaveletFormerNet brings pleasant results in subjective visualization, we can still see from Fig. 17 that some artifacts are introduced when processing images of dense fog. We believe that is due to the limited real-world dense fog data pairs. Real-world data collection is complex, so training on limited data affects the expression capacity of the data-driven models. Therefore, this is our subsequent work to improve our model and introduce more and better quality real-world datasets in the future.

## 7. Acknowledgements

This work was partly supported by the Applied Basic Research Project of Department of Science & Technology of Liaoning province under Grant 2022JH2/101300274 and partly by the Educational Department of Liaoning Province under Grant No.LJKMZ20220679.

## CRediT authorship contribution statement

**Shengli Zhang:** Conceptualization of this study, Methodology, Software, Writing - Original Draft. **Zhiyong Tao:** Supervision, Writing - Review Editing. **Sen Lin:** Supervision, Writing - Review Editing.